%% file: main.tex
\documentclass[sn-mathphys-num]{sn-jnl}%

\usepackage{graphicx}%
\usepackage{multirow}%
\usepackage{amsmath,amssymb,amsfonts}%
\usepackage{amsthm}%
\usepackage{mathrsfs}%
\usepackage{xcolor}%
\usepackage{textcomp}%
\usepackage{manyfoot}%
\usepackage{booktabs}%
\usepackage{algorithm}%
\usepackage{algorithmicx}%
\usepackage{algpseudocode}%
\usepackage{listings}%
\usepackage{mathtools}
\usepackage{tabularx}

\usepackage[nolist]{acronym}
\usepackage{tikz}
\usetikzlibrary{shapes,decorations,arrows,calc,arrows.meta,fit,positioning}
\tikzset{
    -Latex,auto,node distance =1 cm and 1 cm,semithick,
    state/.style ={circle, draw, minimum width = 0.7 cm},
}
\usepackage{caption}
\usepackage{subcaption}
\usepackage{thm-restate}

\usepackage{siunitx}
\newtheorem{assumption}{Assumption}

\newif\ifreview
\reviewfalse
\ifreview
    \usepackage[mathlines]{lineno}
    \AtBeginDocument{\linenumbers}
\fi

\include{symbols}

\begin{acronym}
    \acro{ERM}{Empirical Risk Minimization}
    \acro{i.i.d.}{independent and identically distributed}
    \acro{IRM}{Invariant Risk Minimization}
\end{acronym}

\begin{document}

\title{
    Invariance Pair Guidance: Robustness to Spurious Correlations via Corrective Gradients
}

\author*[1,2]{\fnm{Martin} \sur{Surner}}
\email{Martin.Surner@haw-landshut.de}

\author[1,2]{\fnm{Abdelmajid} \sur{Khelil}}
\email{Abdelmajid.Khelil@haw-landshut.de}

\author[3,4]{\fnm{Ludwig} \sur{Bothmann}}
\email{Ludwig.Bothmann@lmu.de}

\affil*[1]{\orgdiv{Institute for Data and Process Science}, \orgname{Landshut University of Applied Sciences}, %
\orgaddress{\city{Landshut}, \country{Germany}}}

\affil[2]{\orgdiv{Computer Science Department}, \orgname{Landshut University of Applied Sciences}, %
\orgaddress{\city{Landshut}, \country{Germany}}}

\affil[3]{\orgdiv{Department of Statistics}, \orgname{LMU Munich}, %
\orgaddress{\city{Munich}, \country{Germany}}}
\affil[4]{\orgdiv{Munich
Center for Machine Learning (MCML)}, \orgaddress{\city{Munich}, \country{Germany}}
}

\abstract{
Machine learning models are inherently bound to the distribution of the training data, often exploiting non-causal shortcuts. 
As a result, achieving robustness to spurious correlations remains a challenge.
While existing approaches rely on data manipulation or re-weighting strategies to achieve robustness, they typically require dense group labels, multiple training domains, or specialized pre-processing.
We propose Invariance Pair Guidance (IPG), a method to mitigate reliance on spurious correlations using a sparse set of counterfactual pairs. 
Unlike other methods demanding extensive supervision, IPG utilizes a novel dual-update mechanism to dynamically correct the optimization trajectory.
We generate input pairs that isolate the spurious attribute to define the invariance, a characteristic that should not affect the outcome of the model. 
Based on these pairs, we define a corrective gradient that complements the traditional gradient descent approach.
The correction adapts via a predefined invariance condition. 
Experiments on ColoredMNIST, Waterbirds-100, and CelebA datasets demonstrate the effectiveness of our approach and its robustness to group shifts, supported by a theoretical convergence analysis. 
IPG offers a data-efficient and theoretically grounded path to robustness.
}

\keywords{Machine Learning, Robustness, Gradient Manipulation, Spurious Correlations, Representation Learning}

\maketitle 
\begingroup
\let\thefootnote\relax\footnotetext{This version of the article has been accepted for publication, after peer review (when applicable) but is not the Version of Record and does not reflect post-acceptance improvements, or any corrections. The Version of Record is available online at: https://doi.org/10.1007/s10994-026-07092-0}
\endgroup

\section{Introduction}
\label{sec:introduction}
\newsavebox{\leftimagebox}
\begin{figure}
    \centering
    \begin{minipage}[t]{0.285\textwidth}
        \vspace{0pt}
        \begin{subfigure}[t]{\textwidth}
            \includegraphics[width=\textwidth]{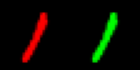}
            \caption{}
            \label{fig:colored_mnist_invariance_pair}
        \end{subfigure}%
        \vspace{2mm}%
        
        \begin{subfigure}[t]{\textwidth}%
            \includegraphics[width=\textwidth]{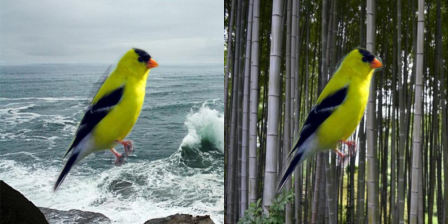}%
            \caption{}%
            \label{fig:waterbird_invariance_pair}
        \end{subfigure}%
        \vspace{2mm}%
        
        \begin{subfigure}[t]{\textwidth}
            \includegraphics[width=\textwidth]{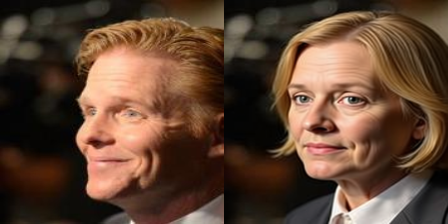}
            \caption{}
            \label{fig:celeba_invariance_pair}
        \end{subfigure}
    \end{minipage}
    \hfill
    \begin{minipage}[t]{0.705\textwidth}
        \vspace{0pt}
        \begin{subfigure}[b]{\textwidth}
            \includegraphics[width=\textwidth, height=7.67cm,
            keepaspectratio]{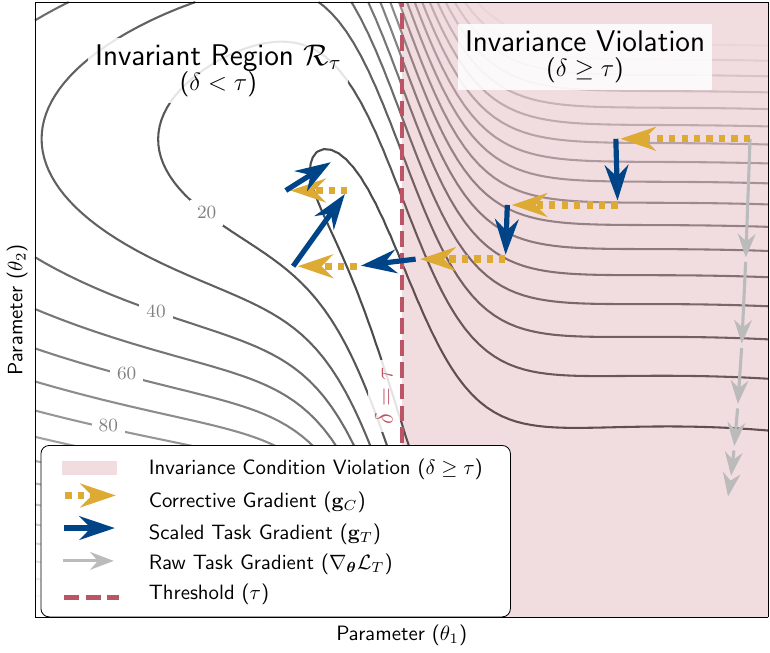}
        \caption{}
        \label{fig:grad_corr}
        \end{subfigure}
    \end{minipage}
    \caption{
    Invariance pairs for (a) ColoredMNIST, (b) Waterbirds-100, and (c) CelebA define the spurious correlations by contrast (color, background, gender, respectively). 
    The model must remain invariant to these features during the digit, bird type, and hair color classification tasks.
    The pairs are used to formulate the invariance condition and corrective gradient.
    (d) Schematic visualization of our approach in model space. 
    The parameter $\theta_1$ activates the spuriously correlated feature, while $\theta_2$ captures the causal feature. 
    The first three task gradients (blue) are scaled to two-thirds the length of the corrective gradient due to the violation of the invariance condition.
    }
    \label{fig:comparison}
\end{figure}

The ability to learn representations from data makes neural networks highly applicable to a wide range of tasks.
However, standard training assumes that training and test data are \ac{i.i.d.}, an assumption that rarely holds in real-world scenarios.
For instance, in autonomous driving, models inevitably encounter unknown situations or shifted distributions \cite{kohli_enabling_2020}.
To operate safely in such environments, models must generalize to unseen data by learning robust, causal features rather than relying on dataset-specific correlations \cite{wang_generalizing_2023}.

A key barrier to robustness is the model's tendency to internalize spurious correlations, features that correlate with the label in the training set but are not causally predictive.
Examples are widespread: animal detection models often rely on background scenery (e.g., snow vs. grass) \cite{beery_recognition_2018, bothmann_automated_2023}, and medical diagnostic models have been shown to rely on hospital-specific tags on radiographs rather than pathology \cite{degrave_ai_2021}.
Similarly, substantial disparities in gender classification accuracy arise when models rely on skin color rather than facial features \cite{buolamwini_gender_2018}.
If these spurious characteristics are not explicitly suppressed, models fail when the spurious correlation breaks in the unseen environment \cite{wimmer_trust_2025}.

To mitigate this risk, we aim to separate the defining characteristics of a class from spurious attributes.
To this end, we introduce the concept of \textit{invariance pairs} \cite{bai_invariant_2025}.
While conceptually related to contrastive learning \cite{le-khac_contrastive_2020}, IPG differs fundamentally in efficiency and scope. 
Instead of structuring the entire latent space via dense, random augmentations, IPG utilizes a sparse set of targeted pairs (often $<3\%$ of the training data) to efficiently correct specific spurious dependencies via a dynamic gradient update.
An invariance pair consists of two samples that share the same semantic content (e.g., the same digit) but differ in a specific spurious attribute (e.g., color) (Fig.~\ref{fig:colored_mnist_invariance_pair}).
We define \textit{invariance} as the requirement that the model's internal representation should remain identical for both elements of the pair.
This creates a direct metric for robustness: if the model's output varies within a pair, it has internalized a spurious dependency.

While identifying spurious attributes is non-trivial, recent advancements in generative modeling provide new possibilities in this process.
In many high-stakes domains, experts already possess the domain knowledge of what should be invariant (e.g., a tumor classification should be invariant to the hospital scanner ID).
Additionally, there are methods to identify such dependencies \cite{abid_meaningfully_2022, goyal_counterfactual_2019}.
We also utilize generative AI (e.g., instruction-based image editing) to synthesize counterfactual invariance pairs, such as gender-swapped portraits, which are otherwise difficult to obtain.
In this work, we assume that a sparse set (e.g., $3\%$ of the training data) of such pairs is available, either from domain knowledge or generative augmentation, and focus on the optimization challenge: how to efficiently utilize them to correct the model's behavior.

We introduce \underline{I}nvariance \underline{P}air \underline{G}uidance (IPG), a method that explicitly enforces these invariances during training.
Unlike standard regularization, IPG extends stochastic gradient descent with a \textit{corrective gradient} step, inspired by \cite{van_baelen_constraint_2022}.
This correction minimizes the discrepancy between the learned representations of invariance pairs.
Crucially, we use an adaptive scaling mechanism based on an \textit{invariance condition}: the corrective gradient is dynamically prioritized only when the model demonstrates a violation of the invariance (i.e., high disagreement between pairs).
This allows the model to internalize the invariance while optimizing on the primary task.

We evaluate IPG on three datasets representing synthetic, composite, and real-world shifts: ColoredMNIST, Waterbirds-100, and CelebA.
Our key contribution is a novel dual-update optimization mechanism that utilizes invariance pairs to improve robustness against shifts in both perfect and partial spurious correlations, even under label noise. 
Figure~\ref{fig:comparison} illustrates the approach and invariance pairs for each dataset.
\footnote{Code is available at \href{https://github.com/msurner/ipg}{https://github.com/msurner/ipg}.}

The remainder of the paper is structured as follows. 
Section~\ref{sec:related_work} discusses related work. 
In Section~\ref{sec:methodology}, we present IPG.
Section~\ref{sec:experiments} evaluates the performance, 
Section~\ref{sec:conclusion} concludes. 

\section{Related Work}
\label{sec:related_work}

We categorize approaches for robustness and out-of-distribution generalization into three streams: data manipulation, representation learning, and learning strategies.
Our proposed IPG method sits at the intersection, integrating explicit representation guidance (invariance pairs) with an adaptive learning strategy (dual-update scheme).

\paragraph{Data Manipulation.}
These methods aim to bridge the gap between training and target distributions via augmentation or generation.
Standard techniques vary image style \cite{wang_learning_2021} or use synthetic data \cite{prakash_structured_2019}.
Perturbation-based strategies, such as adversarial augmentation \cite{huang_robustness_2023, puli_nuisances_2024}, generate worst-group perturbations to balance the data; for instance, DAIR \cite{huang_robustness_2023} regularizes models against such variations.
Other approaches enrich the data with concept-level information, utilizing pseudo-labels \cite{nam_spread_2022} or concept banks \cite{koh_concept_2020}.
A distinct sub-category aims to subtract spurious information via pre-processing.
Shock Graph \cite{narayanan_shape-biased_2021} explicitly converts images into shape-based graph representations, thereby discarding texture and color information entirely.
Similarly, Noisy Counterfactual Matching (NCM) \cite{bai_invariant_2025}, identifies a linear spurious subspace from counterfactual pairs and projects the input data orthogonally to remove these directions.
\textit{Distinction:} IPG differs from these approaches in three key ways: 
(i) unlike augmentation methods (DAIR) that require generating synthetic samples for the entire dataset, IPG guides training using a sparse set of pairs; 
(ii) unlike concept banks, it avoids dense annotation costs; and 
(iii) unlike feature removal (Shock Graph/NCM), IPG preserves full input information, teaching the model to dynamically ignore spurious correlations rather than statically deleting them.

\paragraph{Representation Learning.}
The goal of these methods is to disentangle features or learn domain-invariant representations.
Domain-invariant learning techniques, such as \ac{IRM} \cite{arjovsky_invariant_2019}, enforce an optimal classifier across varying training domains.
A related subfield, feature disentanglement, aims to separate latent factors using generative models \cite{choi_robustnet_2021} or causal scaffolding \cite{sun_recovering_2021}.
Further, EPG \cite{rao_studying_2023}, and GALS \cite{petryk_guiding_2022} offer explicit guidance.
EPG penalizes feature attributions (calculated via B-cos or IxG) that fall outside a target bounding box, thereby effectively guiding the model spatially.
GALS similarly aligns model attention with language-derived ground truths.
DKT \cite{zhou_robustness_2025} generates counterfactual features by transferring semantic statistics between spuriously correlated groups to break dependencies.
\textit{Distinction:} While EPG and GALS require bounding boxes or language encoders to define spatial attention guidance, IPG defines semantic guidance via comparative pairs, making it applicable even when spatial localization is difficult. 
In contrast to IPG, DKT relies on group labels.

\paragraph{Learning Strategies.}
Robustness can also be enforced through specialized optimization objectives.
GroupDRO \cite{sagawa_distributionally_2020} minimizes the loss of the worst-performing group, providing strong robustness guarantees if group labels are known.
In the absence of such labels, Identify-then-Mitigate approaches like JTT~\cite{liu_just_2021} and EvA-E~\cite{he_erasing_2025} first utilize a biased model to identify misclassified samples, then up-weight them (JTT) or prune high-evidence activations (EvA-E).
Similarly, LfF~\cite{nam_learning_2020} employs a dual-model architecture to simultaneously amplify and counteract bias.
Gradient-based methods also play a role; Fishr~\cite{rame_fishr_2022} enforces domain-invariance by matching the variances of gradients across domains, adapting the covariance-alignment principle of CORAL \cite{sun_return_2016}.
\textit{Distinction:} Unlike these methods, IPG operates on a single domain without requiring group labels (vs. GroupDRO), multiple domains (vs. Fishr), or the presence of bias-conflicting samples in the training set (vs. JTT/LfF/EvA-E). 

\section{IPG: Invariance Pair Guidance} 
\label{sec:methodology}
In this section, we describe our approach to formulating the invariance pairs, the corrective gradient, and the adaptive scaling by the invariance condition.

\subsection{Preliminaries and Overview}
\label{sec:preliminaries}

We consider a classification task with input space $\inputspace$ and label space $\mathcal{Y} = \{1, \dots, K\}$.
A training set $\trainDomain \coloneq \{(\x_j, y_j)\}^n_{j=1}$ is drawn from a joint distribution $P_{\trainDomain}$.
Associated with each instance $\x_j$ is a single spurious attribute $a_j \in \mathcal{A} = \{0,1\}$.
While $a_j$ is causally irrelevant to the label $y_j$, the training distribution typically exhibits a strong statistical correlation between specific values of $a$ and $y$.
We define a \textit{group} $g$ as unique combination of the attribute and label:
$g \coloneq (a, y) \in \mathcal{A} \times \mathcal{Y}$. 
Standard models inherently favor correlational patterns over causal ones, frequently relying on the spurious correlation between $a$ and $y$ when amplified by dataset bias or simplicity bias.
In extreme cases, certain groups may be entirely absent from $\trainDomain$ (perfect spurious correlation).

Our goal is to learn a model $f$ that generalizes well to a test distribution $P_{\testDomain}$ where the correlation between $a$ or $y$ is reduced or inverted ($P_{\trainDomain} \neq P_{\testDomain}$).
To measure robustness across all subpopulations, we target the \textit{worst-group accuracy} \cite{sagawa_distributionally_2020}:

\begin{equation*}
    \accWg(f) \coloneq \min_{g \in \mathcal{A} \times \mathcal{Y}} \mathbb{E}_{(\x, y) \sim P_{\testDomain^g }}[\ind_{y = f(\x)}].
\end{equation*}
where $P_{\testDomain^g}$ denotes the distribution of test samples belonging to group $g$.

\paragraph{Assumptions.}
Our method relies on a set of \textit{invariance pairs} $\invpairs$.
We assume that for any pair $(\invX, \invX') \in \invpairs$, both samples share the same semantic label $y$ and features, but differ in their spurious attribute ($a \neq a'$).
These pairs serve as explicit examples of the invariance the model must learn.

\begin{figure}[htbp]
    \centering
    \includegraphics[width=.7\linewidth]{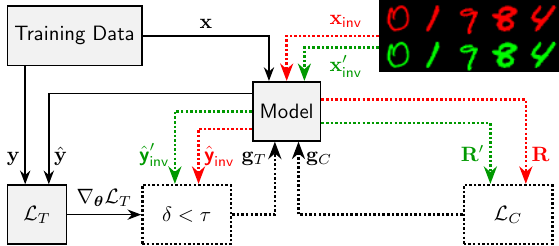}
    \caption{
    Schematic overview of the IPG training method (dotted) as an extension to the traditional approach using the example of ColoredMNIST. 
    }
    \label{fig:system_model}
\end{figure}

\paragraph{Method Overview.}
We propose an extension to stochastic mini-batch gradient descent \cite{dekel_optimal_2012} that enforces group robustness through a dual-update mechanism (Fig.~\ref{fig:system_model}).
First, utilizing batches sampled from $\invpairs$, we compute a \textit{corrective gradient} $\gradCorr$ that minimizes the discrepancy between the internal representations of invariant pairs.
Second, we perform a task update using the standard loss gradient, but with a critical modification: we monitor a \textit{disagreement rate} $\disagree$ between pair predictions.
If $\disagree$ exceeds a threshold $\thresh$, indicating reliance on spurious features, we strictly scale the task gradient to prioritize the corrective signal.
This adaptive scaling allows the model to learn invariance when necessary, while retaining optimization capacity for the task loss when the invariance condition is satisfied.
We detail these components in the following section.

\subsection{Rationale and Invariance Pair Definition}
To compare learned characteristics and formulate the corrective gradient $\gradCorr$, we utilize a latent representation known as the \textit{rationale} matrix \cite{chen_domain_2023}.
The rationale explicitly encodes the contribution of learned features to the final classification.
We decompose the model $f$ into two modules: a feature extractor $\extractor$, mapping inputs $\x$ to features $\z \coloneq \extractor(\x)$, and a classifier $h$, mapping $\z$ to logits $\mathbf{o} \coloneq h(\z)$.
Specifically, the classifier is defined by $h(\z) := \z\params_h^T + \params_{h,bias}$, where $\params_h\in \mathbb{R}^{K\times D}, \params_{h,bias}\in \mathbb{R}^{K}$ are the classifier weights and bias, respectively.
The rationale $\R$ is defined as the element-wise product of the feature activations and $\params_h$:
\begin{equation*}
    \R \coloneq \diag(\z)\params_h^T \in \mathbb{R}^{D\times K},
\end{equation*}
where $D$ and $K$ denote the feature dimension and number of classes, respectively.
Unlike raw logits, $\R$ retains fine-grained semantic information by incorporating the final layer's weights, making it an ideal candidate for guiding the training process.

To identify and penalize reliance on spurious correlations, we contrast rationales derived from a set of \textit{invariance pairs} $\invpairs$.
We use a set of invariant pairs $(\invX, \invX') \in \invpairs$ to define a specific invariance.
Each pair $(\invX, \invX') \in \invpairs$ contains two variations of the same semantic content that differ only in a specific spurious attribute.
Ideally, since the semantic label $y$ and class-relevant features are identical, the classification, and thus the rationale, should remain invariant to this change.
For example, in the ColoredMNIST dataset, a pair consists of the same digit rendered in different colors (Fig.~\ref{fig:colored_mnist_invariance_pair}).
These pairs provide the necessary ground truth to define the target model behavior independent of the spurious correlation.

\subsection{Corrective Gradient and Invariance Condition}

To guide the neural network towards robust representations, we use a sequential optimization strategy.
During training, we sample a batch of invariance pairs $\invpairsbatch$ from $\invpairs$ to compute the corrective updates.
We establish a distance measure to quantify the internalization of the specified invariance.
For a given invariance pair $(\invX, \invX') \in \invpairsbatch$, let $\R, \R' \in \mathbb{R}^{D \times K}$ denote their corresponding rationale matrices.
The model predictions $\invpairpred$ are equal to an aggregation of these rationales and an input-agnostic bias $\params_{h,bias} \in \mathbb{R}^{K}$, which is then passed to the $\softmax$ function:
\begin{equation}
    \invpairpred = \softmax\left( \underbrace{\mathbf{1}_D^T \R + \params_{h,bias}}_{=\mathbf{o} = h(\z)} \right), %
    \label{eq:rationale_prediction}
\end{equation}
where $\mathbf{1}_D$ is a $D$-dimensional column vector of ones. 

Since the pairs are invariant by definition, their rationales, and consequently their predictions, should be aligned.
We define an alignment loss based on the mean pairwise Frobenius distance:
\begin{equation*}
    \lossAlign \coloneq \frac{1}{\vert\invpairsbatch\vert}\sum_i\Vert \R_i - \R_i'\Vert_F.
\end{equation*}
To prevent model collapse, where the model maps all inputs to a trivial constant representation, we enforce a uniformity constraint.
We define a hinge loss to ensure that rationales for class-conflicting elements differ by at least a fixed margin $m$:
\begin{equation*}
    \lossUniform \coloneq \frac{1}{2|\invpairsbatch|^2} \sum_{\mathbf{Q} \in \{\mathbf{R}, \mathbf{R}'\}} \sum_{i,j} \ind(y_i \neq y_j) \cdot \max \left(0, \margin - \Vert \mathbf{Q}_i - \mathbf{Q}_j \Vert_F\right).
\end{equation*}
The total corrective loss is the sum of these components: $\lossCorr \coloneq \lossAlign + \lossUniform$.

We extend standard mini-batch gradient descent by performing two distinct update steps per iteration: a \textit{correction step} followed by a \textit{scaled task optimization step}.

\paragraph{Step 1: Correction.}
We first compute the corrective gradient $\gradCorr \coloneq \nabla_{\params}\lossCorr$.
This gradient directs the model parameters $\params$ towards a region where the invariance constraints (alignment and uniformity) are satisfied.
The optimizer performs the first update using $\gradCorr$.

\paragraph{Step 2: Scaled Task Optimization.}
Subsequently, we update the model to minimize the task loss $\lossTask$, defined as cross-entropy loss.
However, we do not apply the raw gradient $\nabla_{\params} \lossTask$ directly. 
Instead, we compute a scaled task gradient $\gradTask$ that respects the invariance established in the previous step.
We quantify this dependency using the disagreement rate $\disagree$, defined as the fraction of invariant pairs yielding inconsistent class predictions:
\begin{equation*}
    \disagree \coloneq \frac{1}{|\invpairsbatch|} \sum_{k} \ind(\arg\max \invpairpred[,k] \neq \arg\max \invpairpred[,k]').
\end{equation*}

To formalize the switching behavior of our optimizer, we define an invariance condition: $\disagree < \thresh$, based on a fixed predefined threshold $\thresh \in [0,1]$.
When $\disagree \ge \thresh$, the model violates the invariance condition, triggering a strict corrective mechanism.
The mechanism rescales the raw task gradient $\nabla_{\params}\lossTask$ to a fixed fraction $\stepSize \in \left[0,1\right)$ of the magnitude of the corrective gradient $\gradCorr$.
Otherwise, gradient magnitudes are harmonized.
We calculate the final scaled task gradient $\gradTask$ based on the invariance condition:
\begin{equation}
    \gradTask \coloneq \begin{cases}
        \stepSize \cdot \frac{\nabla_{\params} \lossTask}{\|\nabla_{\params} \lossTask\|_2} \|\gradCorr\|_2 & \text{if } \disagree \ge \thresh \\[2ex]
        \text{clip}\big(\nabla_{\params} \lossTask, \, 2\cdot\|\gradCorr\|_2\big) & \text{else},
    \end{cases}
    \label{eq:loss_grad}
\end{equation}
where $\text{clip}(\mathbf{u}, c)$ rescales vector $\mathbf{u}$ such that $\|\mathbf{u}\|_2 \le c$, maintaining its direction.
Finally, the optimizer performs the second update using $\gradTask$.
We update the weights using an optimizer that can involve momentum-based techniques, allowing for a stabilizing effect. 
This adaptive scaling mechanism prioritizes invariant representation learning when spurious correlations are detected, while allowing the task loss to dominate when the invariance condition is satisfied.

An overview of the presented method is shown in Alg. \ref{alg:approach} defining the extended update function for one gradient descent step for a training-data batch given by $(\batchX, \batchY)$.

\begin{algorithm}
\caption{IPG update step}
\label{alg:approach}
\begin{algorithmic}[1]
\Require Model weights at step $t$ $\params^{(t)}$
\Require Training data batches $\batchX, \batchY$, invariance pairs batch $\invpairsbatch$
\Require Optimization hyperparameters: learning rate $\lr$
\Require Algorithm parameters: disagreement rate threshold $\thresh$, step size $\stepSize$, minimal gradient length $\varepsilon$, uniformity margin $m$

\State $\invX, \invpairpred, \invX', \invpairpred' \leftarrow \invpairsbatch$ \Comment{Unpack batch}
\State $\z, \z' \leftarrow \extractor(\invX; \params^{(t)}_\extractor), \extractor(\invX'; \params^{(t)}_\extractor)$ 
\State $\R, \R' \leftarrow \diag(\z)(\params^{(t)}_h)^T, \diag(\z')(\params^{(t)}_h)^T$ \Comment{Compute rationale matrices $\in \mathbb{R}^{D \times K}$}
\State $\invpairpred, \invpairpred' \leftarrow \softmax(h(\z; \params^{(t)}_h)), \softmax(h(\z'; \params^{(t)}_h))$

\State $\lossAlign \leftarrow \frac{1}{\vert\invpairsbatch\vert}\sum_i\Vert \R_i - \R_i'\Vert_F$

\State $\lossUniform \leftarrow \frac{1}{2|\invpairsbatch|^2} \sum_{\mathbf{Q} \in \{\mathbf{R}, \mathbf{R}'\}} \sum_{i,j} \ind(y_i \neq y_j) \cdot \max \left(0, \margin - \Vert \mathbf{Q}_i - \mathbf{Q}_j \Vert_F\right)$

\State $\lossCorr \leftarrow \lossAlign + \lossUniform$

\State $\disagree \leftarrow \frac{1}{|\invpairsbatch|} \sum_{k} \ind(\arg\max \invpairpred[,k] \neq \arg\max \invpairpred[,k]')$ \Comment{Compute disagreement rate}

\State $\gradCorr \leftarrow \nabla_{\params^{(t)}} \lossCorr$ 

\State $\params^{(t+1)} \leftarrow \update(\params^{(t)}, \lr,\gradCorr)$
\Comment{Apply corrective step}

\State $\hat{\y} \leftarrow \softmax(h(\extractor(\x; \params^{(t+1)}_\extractor);\params^{(t+1)}_h))$ %

\State $\gradTask \leftarrow \nabla_{\params^{(t+1)}} \loss(\y, \hat{\y})$ %

\If{$\disagree \ge \thresh$} \Comment{Invariance condition}
    \State $\gradTask \leftarrow \stepSize \cdot \frac{\gradTask}{||\gradTask||_2} || \gradCorr ||_2$
\Else
    \If{$||\gradTask||_2> 2 \cdot \max\{\varepsilon, || \gradCorr ||_2\}$}
        \State $\gradTask \leftarrow 2 \cdot \frac{\gradTask}{||\gradTask||_2}\max\{\varepsilon, || \gradCorr ||_2\}$
    \EndIf
\EndIf

\State $\params^{(t+2)} \leftarrow \update(\params^{(t+1)}, \lr,\gradTask)$ \Comment{Apply scaled task gradient} %

\end{algorithmic}
\end{algorithm}

\section{Theoretical Analysis}
\label{sec:theoretical_analysis}
In this section, we analyze the theoretical properties of the approach, focusing on two aspects: the monotonic correction, its resulting convergence properties, and a comparative analysis of computational complexity.

\subsection{Monotonic Correction and Convergence Properties}
For this analysis, we show that IPG results in a monotonic descent on $\lossCorrParams{}$ when the invariance condition is violated.
Based on the theorem, we further derive convergence properties.
We introduce the following simplifications:
First, the parameters $\params_k$ at the start of the IPG update step (Alg. \ref{alg:approach}) are fixed.
Second, we treat the optimization process as standard gradient descent (ignoring momentum for the step-wise analysis).
All norms applied are Euclidean. 
We rely on the following common assumptions:
\begin{assumption}[Convexity]\label{ass:convexity}
The function $\lossCorrParams{}$ is convex over the domain $\paramsDomain \subseteq \mathbb{R}^d$.\end{assumption}
\begin{assumption}[Smoothness]\label{ass:smoothness}
The loss function $\lossCorrParams{}$ is $L$-smooth, meaning its gradient is Lipschitz continuous. 
Consequently, for all $\params, \params' \in \mathbb{R}^d$:
$$\lossCorrSub{, \params'} \leq \lossCorrSub{, \params} + \langle \nabla \lossCorrSub{, \params}, \params' - \params \rangle + \frac{L}{2} \|\params' - \params\|^2.$$
\end{assumption}
\begin{assumption}[Joint Feasibility]\label{ass:joint_feasibility}
There exists a non-empty subset of parameters $\paramsDomain^* \subset \paramsDomain$ such that for $\params \in \paramsDomain^*$, $\lossAlignSub{, \params} = 0$ and $\lossUniformSub{, \params} = 0$. 
This implies the rationale dimension $D$ is sufficient to satisfy margin constraints without violating invariance.
\end{assumption}
\begin{restatable}[Monotonic Descent of $\lossCorrParams{}$]{theorem}{monotonicDescent}
\label{thm:monotonic_desc}
Let Assumptions~\ref{ass:convexity} and \ref{ass:smoothness} hold.
Suppose the invariance condition is violated ($\disagree \ge \thresh$) and $\thresh \in (0,1]$.
Then, provided the task gradient scaling factor $\stepSize < 1$ and the learning rate $\lr$ is sufficiently small, the algorithm ensures monotonic descent:
$$\lossCorrParams{k+1} < \lossCorrParams{k}.$$
\end{restatable}
The full proof is provided in Appendix~\ref{app:proof}, showing a dominance of $\gradCorr$ when $\stepSize < 1$ and thus the method results in a monotonic descent under given assumptions.

\begin{restatable}[Convergence Properties]{corollary}{convergenceProperties}
\label{cor:convergence_properties}
Let Assumptions \ref{ass:convexity}, \ref{ass:smoothness}, and \ref{ass:joint_feasibility} hold. 
If the optimization follows the Monotonic Descent property (Theorem~\ref{thm:monotonic_desc}) whenever the invariance condition is violated ($\disagree > \thresh$), then, given Eq.~\ref{eq:rationale_prediction}:

\begin{enumerate}
    \item \textbf{Finite Feasibility}: For any relaxed threshold $\tau > 0$, the sequence of parameters $\{\params_k\}$ reaches the invariant region $\invRegion = \{ \params \mid \disagree_{\params} < \thresh \}$ in a finite number of steps.
    \item \textbf{Finite Exact Convergence}: For the strict case $\thresh = 0$, the discrete disagreement rate reaches the exact invariant state $\disagree_{\params} = 0$ in a finite number of steps.
\end{enumerate}
\end{restatable}
The proof is provided in Appendix~\ref{app:proof}, deriving both properties from the Theorem~\ref{thm:monotonic_desc}.

\subsection{Extension to Stochastic and Momentum-based Optimization}
While the theoretical analysis above assumes full-batch gradient descent for clarity, the results generalize to practical training settings under standard assumptions.
\paragraph{Stochastic Gradient Descent (SGD).}
In the stochastic setting, the monotonic descent condition (Theorem~\ref{thm:monotonic_desc}) translates to a \textit{descent in expectation}: 
$\mathbb{E}[\lossCorrParams{k+1}] \leq \lossCorrParams{k}$. 
Provided that the learning rate follows a standard decay schedule and the stochastic gradients satisfy bounded variance assumptions, the convergence to the invariant region $\invRegion$ holds almost surely.
Specifically, for $\tau > 0$, the parameter trajectory enters the invariant region $\invRegion$ with probability 1. 
For $\tau = 0$, the stability margin ensures that the parameters reach the exact invariant state $\disagree_{\params_k} = 0$ in finite time almost surely.

\paragraph{Momentum.}
For momentum-based optimizers, the update vector is a history-dependent aggregation of past gradients.
In this context, the proposed gradient scaling (Eq.~\ref{eq:loss_grad}) ensures that all \textit{new} information entering the momentum buffer satisfies the invariance constraint. 
While the momentum history may introduce a transient lag, the spurious components from previous steps decay exponentially (controlled by the momentum factor $\beta < 1$). 
Since the source term is continuously corrected, the momentum buffer effectively discards spurious alignment, guaranteeing that the update direction aligns with the invariant region or exact invariant state in finite time.

\paragraph{Summary.}
The theoretical analysis demonstrates that the corrective gradient induces a monotonic descent in the correction error $\lossCorr$ whenever the invariance condition is violated.
For relaxed constraints ($\tau > 0$), this guarantees that the model satisfies the condition in a finite number of steps.
For strict constraints ($\tau = 0$), the \textit{discrete} classification becomes invariant in finite time once the correction error falls below the decision boundary's stability margin.

\subsection{Computational Complexity Analysis}
The computational complexity of IPG during the training phase depends on the number of epochs $N$, the dataset size $D$, the task batch size $B_D$, and the invariance pair batch size $B_I$.
We define the complexity in terms of the number of forward and backward passes through the network. 
Standard \ac{ERM} requires one forward and one backward pass per sample, scaling linearly as $\mathcal{O}(N \cdot D)$.

In contrast, IPG introduces an overhead due to the corrective mechanism.
For every task batch, we sample a batch of invariant pairs (size $B_I$) and perform a corrective update.
Since pairs consist of two images $(\invX, \invX')$, this adds $2 \cdot B_I$ forward passes and a corresponding backward pass per iteration.
The total number of iterations per epoch is $\lceil D/B_D \rceil$.
Thus, the total operations for IPG can be approximated as:
\begin{equation*}
    C_{\text{IPG}} \approx N \cdot \left( D + \left\lceil\frac{D}{B_D}\right\rceil \cdot 2 \cdot B_I \right) \approx N \cdot D \left( 1 + \frac{2 B_I}{B_D} \right).
\end{equation*}
While the asymptotic complexity remains linear, $\mathcal{O}(N \cdot D)$, the computational complexity of IPG scales linearly as $N \cdot D (1 + 2B_I / B_D)$. 
In our experiments, we set  $B_I = B_D$.
However, the results in Appendix~\ref{app:sensitivity} indicate that accuracy remains comparable at a ratio of $B_I / B_D = 0.25$. 

We compare the computational characteristics of IPG with related approaches in Table~\ref{tab:computational_complexity}.
We categorize methods based on their estimated relative runtime factor compared to standard ERM.

\begin{table}[ht]
\centering
\caption{Runtime cost comparison. The relative cost estimates the multiplier of training time relative to ERM (where ERM $= 1\times$).}
\label{tab:computational_complexity}
\begin{tabular}{l l c}
    \toprule
    Approach & Mechanism & Relative Cost ($\approx$) \\
    \midrule
    ERM & --  & $1\times$ \\
    \midrule
    \multicolumn{3}{l}{\textit{Pre-processing / Input Transformation}} \\
    Shock Graph \cite{narayanan_shape-biased_2021} & Input Transformation & $1\times$ (+ Pre.) \\
    NCM \cite{bai_invariant_2025} & Linear Projection &  $1\times$ (+ Pre.) \\
    \midrule
    \multicolumn{3}{l}{\textit{Explanation-Guided}} \\
    EPG \cite{rao_studying_2023} & Input Gradient Penalty  & $2\times$ \\
    GALS \cite{petryk_guiding_2022} & Attention Alignment  & $2\times$ \\
    \midrule
    \multicolumn{3}{l}{\textit{Penalty-based Methods}} \\
    IRM \cite{arjovsky_invariant_2019} & Gradient Norm Penalty  & $2\times$ \\
    Fishr \cite{rame_fishr_2022} & Gradient Variance Matching  & $2\times$ \\
    GroupDRO \cite{sagawa_distributionally_2020} & Group-wise Loss  & $1\times$ \\
    \midrule
    \multicolumn{3}{l}{\textit{Two-Stage / Dual-Model}} \\
    JTT \cite{liu_just_2021} & Two-stage Training  & $2\times$ \\
    LfF \cite{nam_learning_2020} & Bias-Amplified Training  & $2\times$ \\
    \midrule
    \multicolumn{3}{l}{\textit{Post-hoc / Feature Editing}} \\
    EvA-E \cite{he_erasing_2025} & Feature Deletion & $1\times$ (+ Post.)$^\dagger$ \\
    \midrule
    \multicolumn{3}{l}{\textit{Proposed Method}} \\
    IPG (Ours) & Dual-Update Mechanism  & $1 + \frac{2B_I}{B_D}\times$ \\
    \bottomrule
\end{tabular}
\vspace{1ex}
{\small $^\dagger$ EvA-E requires a fully trained ERM model ($1\times$) as input. The post-hoc editing step itself is computationally negligible \cite{he_erasing_2025}.}
\end{table}

\section{Experiments}
\label{sec:experiments}

In this section, we present the experimental setup and results along with a detailed discussion.

\subsection{Experimental Setup}
We evaluate IPG on three datasets representing unique challenges in spurious correlation: label noise, perfect spurious correlation, and feature entanglement. 
For baseline comparisons, we report results directly from the original publications. 
We maintain identical experimental protocols (architecture and data splits) to the baselines, with one necessary exception: for ColoredMNIST, we introduce a dedicated validation split partitioned from the test set to conduct strictly isolated hyperparameter tuning, thereby preventing data leakage. 
Full implementation details are provided in Appendix~\ref{app:implementation}.
We report each mean and standard deviation based on $10$ trials.

\paragraph{Label Noise (ColoredMNIST).}
We use ColoredMNIST~\cite{arjovsky_invariant_2019} primarily to study spurious color correlations, explicitly noting that its standard formulation inherently includes 25\% label noise to simulate noisy supervision.
This dataset is structured as a domain generalization benchmark where the task is binary digit classification.
The digit color (red or green) serves as a strongly correlated spurious attribute in the training set ($p_\text{tr} \in \{0.1, 0.2\}$) but is reversed in the test set ($p_\text{te} = 0.9$).
To guide the model, we create the dataset with invariance pairs consisting of the same digit instance recolored with the opposite color (e.g., a red '1' paired with a green '1', see Fig.~\ref{fig:colored_mnist_invariance_pair}).
We employ a standard CNN architecture following the DomainBed framework~\cite{gulrajani_search_2021} and report the mean test accuracy on the out-of-distribution test set.

\paragraph{Perfect Spurious Correlation (Waterbirds-100).}
To evaluate robustness without natural counter-examples, we use the Waterbirds-100 variant, following the protocol of Petryk et al.~\cite{petryk_guiding_2022}.
Unlike the standard benchmark, all samples in the training set contain an attribute that is perfectly correlated with the labels, while being 100\% spurious.
To overcome the lack of support of bias-conflicting samples, we generate invariance pairs by explicitly swapping the background of a given bird instance (e.g., pasting a waterbird onto a land background, see Fig.~\ref{fig:waterbird_invariance_pair}).
This setting tests the model's ability to extrapolate solely through this sparse guidance, as the test dataset consists of samples from all four groups.
We also evaluate the reversed setting, where the target label is the background and the spurious attribute is the bird type.
We train a ResNet-50 backbone~\cite{he_deep_2016} and report worst-group accuracy ($\accWg$) for both the standard and reversed settings.

\paragraph{Feature Entanglement (CelebA).}
We use the CelebA dataset~\cite{sagawa_distributionally_2020} to assess performance in a setting where spurious and semantic features are inherently entangled.
The task is classifying hair color (blond vs. non-blond), which is spuriously correlated with gender.
Unlike the previous datasets where attributes can be perfectly isolated (e.g., via segmentation), real-world attributes such as gender and hair style are semantically linked.
We address this by generating invariance pairs via instruction-based editing using the FLUX.1 Kontext [dev] model~\cite{labs_flux_2025}, prompting it to alter the gender while preserving identity and hair color (Fig.~\ref{fig:celeba_invariance_pair}).
Consequently, the generated invariance pairs inevitably contain feature entanglement and generation artifacts.
See Appendix~\ref{app:pair-generation} for further details.
This setting evaluates whether IPG can extract useful gradients from noisy, entangled pairs using a ResNet-50 model.
We report worst-group accuracy ($\accWg$).

\subsection{Accuracy Comparison}
\label{sec:accuracy_comparison}

\paragraph{ColoredMNIST Results.}
We evaluate IPG on the ColoredMNIST test set, where the spurious correlation (color-label association) is inverted relative to the training set.
As shown in Table~\ref{tab:perf_colored_mnist}, IPG achieves a mean test accuracy of $72.8\%$.
This result effectively recovers the performance level of the grayscale oracle ($73.1\%$) and approaches the theoretical limit of $75\%$ imposed by the dataset's label noise.
In comparison to state-of-the-art methods, IPG is statistically indistinguishable from DIAR-RA \cite{huang_robustness_2023} ($73.1\%$), which relies on color-augmentation and thus breaks spurious correlation.
Notably, IPG outperforms NCM \cite{bai_invariant_2025} ($69.3\%$), a recent approach that also utilizes invariance pairs but uses a static linear subspace projection.
This suggests that IPG's dynamic gradient correction captures non-linear invariances more effectively than NCM's linear removal.
We further observe that IPG outperforms Shock Graph ($71.6\%$), which explicitly discards color information via preprocessing.
Overall, IPG demonstrates superior performance compared to standard invariance baselines (IRM, Fishr) and competes with intensive augmentation strategies like DIAR, validating the effectiveness of sparse invariance guidance in extreme correlation shifts.
Summarizing, IPG reaches oracle-level performance on ColoredMNIST, comparable to the state-of-the-art DIAR-RA.

\begin{table}[ht]
\centering
\caption{Accuracy comparison on ColoredMNIST.}
\label{tab:perf_colored_mnist}
\begin{tabular}{l S[table-format=2.1(1)]}
    \toprule
    Approach & {Mean Acc. ↑} \\
    \midrule
    ERM & \num{16.1 \pm 0.8} \\
    IRM \cite{arjovsky_invariant_2019} & \num{66.9 \pm 2.5} \\
    Fishr \cite{rame_fishr_2022} & \num{68.8 \pm 1.4} \\
    NCM \cite{bai_invariant_2025} & \num{69.3} \\
    meta-IRM \cite{bae_meta-learned_2021} & \num{70.4 \pm 0.9} \\
    Shock Graph \cite{narayanan_shape-biased_2021} & \num{71.6} \\
    DIAR-RA \cite{huang_robustness_2023} & {\bfseries\num{73.1 \pm 0.1}}\\
    IPG (ours) & \num{72.8 \pm 0.5} \\
    \midrule
    Random & \num{50.0 \pm 0.0} \\
    ERM grayscale (oracle) & \num{73.1 \pm 0.4} \\
    Optimal & \num{75.0 \pm 0.0} \\
    \bottomrule
\end{tabular}
\end{table}

\paragraph{Waterbirds-100 Results.}
We further evaluate the performance of IPG on the Waterbirds-100 dataset, which has a perfect spurious correlation in the training set: every waterbird appears against a water background, and every landbird against a land background, with no bias-conflicting samples available.
This creates a challenging zero-shot generalization scenario for the worst-group (e.g., waterbirds on land).
Following the protocol of \citet{rao_studying_2023} and ~\citet{petryk_guiding_2022}, we also examine a reversed setting (Waterbirds-100-reverse), where labels and spurious attributes are swapped (predicting the background class instead of the bird species).

We compare our approach against standard ERM and two state-of-the-art guidance methods: EPG~\cite{rao_studying_2023}, which aggregates attributions from B-cos, $\mathcal{X}$-DNN, and IxG; and the language-guided GALS method~\cite{petryk_guiding_2022}.
Table~\ref{tab:perf_waterbird} summarizes the results. 
In this data-scarce regime, IPG outperforms the baselines with 124 invariance pairs.
In the standard setting, IPG achieves a worst-group accuracy of 86.3\%, outperforming the best baseline (GALS, 56.7\%) by a margin of +29.6 percentage points.
Similarly, in the reverse setting, IPG improves upon GALS by +13.8 percentage points (86.7\% vs. 72.9\%).
We conclude that IPG establishes a new state-of-the-art for this strict spurious correlation setting to the best of our knowledge.
It effectively mitigates the reliance on background features even when the correlation is perfect, by using a sparse set of invariance pairs.

\begin{table}[ht]
\centering
\caption{Accuracy comparison on Waterbirds-100 and Waterbirds-100-reverse.}
\label{tab:perf_waterbird}
\begin{tabular}{l|
    S[table-format=2.1(2)] S[table-format=2.1(2)]|
    S[table-format=2.1(2)] S[table-format=2.1(2)]
}
    \toprule
    & \multicolumn{2}{c|}{Waterbirds-100} & \multicolumn{2}{c}{Waterbirds-100-reverse} \\
    Model & {Mean Acc. ↑} & {$\accWg$ ↑} & {Mean Acc. ↑} & {$\accWg$ ↑} \\
    \midrule
    ERM & \num{69.3 \pm 1.1} & \num{37.9 \pm 1.7} & \num{83.3 \pm 1.8} & \num{62.7 \pm 3.5} \\
    EPG B-cos \cite{rao_studying_2023} & \num{71.1 \pm 0.9} & \num{41.0 \pm 2.1} & \num{83.6 \pm 1.1} & \num{62.8 \pm 2.1} \\
    EPG $\mathcal{X}$-DNN \cite{rao_studying_2023} & \num{73.1 \pm 3.4} & \num{47.0 \pm 9.1} & \num{82.6 \pm 2.0} & \num{63.9 \pm 3.6} \\
    EPG IxG \cite{rao_studying_2023} & \num{78.1 \pm 2.6} & \num{56.1 \pm 7.0} & \num{78.9 \pm 1.9} & \num{56.5 \pm 3.7} \\
    GALS \cite{petryk_guiding_2022} & \num{79.7} & \num{56.7} & \num{86.8} & \num{72.9} \\
    IPG (ours) & {\bfseries\num{89.5 \pm 1.1}} & {\bfseries \num{86.3 \pm 1.9}} & {\bfseries\num{93.0 \pm 0.7}} & {\bfseries \num{86.7 \pm 1.7}} \\
    \bottomrule
\end{tabular}
\end{table}

\paragraph{CelebA Results.}
Finally, we evaluate IPG on CelebA, where the spurious correlation (gender bias) arises from complex, real-world data distributions.
We compare IPG against the upper-bound baseline GroupDRO~\cite{sagawa_distributionally_2020} and DKT~\cite{zhou_robustness_2025} (which uses explicit group labels), and recent label-free approaches including LfF~\cite{nam_learning_2020}, JTT~\cite{liu_just_2021}, and EvA-E~\cite{he_erasing_2025}.
Results are summarized in Table~\ref{tab:perf_celeba}.
IPG ($74.7\%$) achieves a substantial improvement over the standard ERM baseline ($47.8\%$) and outperforms the label-free LfF method ($70.6\%$). 
While IPG lags behind implicit methods like JTT ($81.1\%$) and feature-synthesis methods like DKT ($92.3\%$), it achieves this result using a single-stage training process with sparse input constraints.
Notably, IPG maintains a high mean test accuracy ($90.8\%$), outperforming JTT ($88.0\%$) and LfF ($86.0\%$) in average performance.
This indicates that IPG acts as a conservative regularizer: it successfully mitigates the worst-group failure modes (+26.9 percentage points gain) without over-correcting to the point of degrading the model's core semantic capabilities.

\begin{table}[ht]
\centering
\caption{Accuracy comparison on CelebA.}
\label{tab:perf_celeba}
\begin{tabular}{l 
    S[table-format=2.1(1)]
    S[table-format=2.1(1)]
}
    \toprule
    Model & {Test Acc. ↑} & {$\accWg$ ↑} \\
    \midrule
    ERM & {\bfseries \num{94.9 \pm 0.2}} & \num{47.8 \pm 3.7} \\
    LfF \cite{nam_learning_2020} & \num{86.0} & \num{70.6} \\
    JTT \cite{liu_just_2021} & \num{88.0} & \num{81.1} \\
    GroupDRO \cite{sagawa_distributionally_2020} & \num{92.9 \pm 0.2} & \num{88.9 \pm 2.3} \\
    EvA-E \cite{he_erasing_2025} & \num{88.7} & \num{82.7} \\
    DKT \cite{zhou_robustness_2025} & \num{94.6} & {\bfseries \num{92.3}} \\ 
    IPG (ours) & \num{90.8 \pm 1.6} & \num{74.7 \pm 8.1} \\
    \bottomrule
\end{tabular}
\end{table}

\subsection{Effectiveness and Efficiency Analysis}
\label{sec:effectiveness_sensitivity}
\paragraph{Data Efficiency.} 
We analyze the sensitivity of IPG to the number of invariance pairs (Table~\ref{tab:pair-sensitivity}). 
To this end, we evaluate four exponentially spaced sample sizes ranging from 40 to 1208 pairs.
Remarkably, IPG achieves robust performance with as few as 124 pairs across all datasets. 
Increasing the number of pairs to 1208 pairs leads to diminishing improvements, confirming that IPG relies on a \textit{sparse} signal rather than dense coverage.
This validates our claim of data efficiency compared to contrastive methods requiring massive batch sizes. 
For this reason, we limit our method to $<4\%$ when comparing.

\paragraph{Hyperparameter Stability.} 
We analyze the impact of the corrective step size $\stepSize$ and threshold $\thresh$ (detailed results in Appendix~\ref{app:sensitivity}).
We observe a dichotomy between synthetic and real-world shifts. 
On ColoredMNIST and Waterbirds-100, strict constraints ($\thresh=0.1, \stepSize=0.1$) perform best, effectively mitigating reliance of the model on the color or background bias. 
Conversely, on CelebA, looser constraints ($\thresh \approx  0.5$) prevent the corrective gradient from impeding the learning of core semantic features.

\begin{table}[htb]
    \centering
    \caption{
    Sensitivity to the number of pairs and their percentage share in the training dataset (\% $\trainDomain$).
    }
    \label{tab:pair-sensitivity}

    \begin{tabular}{
        c|
        S[table-format=1.3(3)] 
        S[table-format=1.1] 
        S[table-format=1.3(3)] 
        S[table-format=2.1] 
        S[table-format=1.3(3)] 
        S[table-format=1.2] 
    }
        \toprule
        & \multicolumn{2}{c}{ColoredMNIST} & \multicolumn{2}{c}{Waterbirds-100} & \multicolumn{2}{c}{CelebA} \\
        {Number of Pairs} & {$\accWg$ ↑} & {\% $\trainDomain$} & {$\accWg$ ↑} & {\% $\trainDomain$} & {$\accWg$ ↑} & {\% $\trainDomain$} \\
        \midrule
        40   & \num{67.6 \pm 1.5} & 0.1 & \num{83.9 \pm 1.2} & 0.8 & \num{67.5 \pm 8.5} & 0.02 \\
        124  & \num{71.2 \pm 0.6} & 0.3 & \num{86.3 \pm 1.9} & 2.6 & \num{66.8 \pm 8.7} & 0.08 \\
        388  & \num{72.1 \pm 1.2} & 1.0 & \num{86.9 \pm 0.8} & 8.1 & \num{73.2 \pm 6.2} & 0.2 \\
        1208 & {\bfseries\num{72.8 \pm 0.5}} & 3.2 & {\bfseries\num{87.0 \pm 0.5}} & 25.2 & {\bfseries\num{74.7 \pm 8.1}} & 0.7 \\
        \bottomrule
    \end{tabular}
\end{table}

\subsection{Discussion and Limitations}
\label{sec:discussion}

\paragraph{Oracle Recovery and Label Noise.}
Our experiments on ColoredMNIST demonstrate that IPG effectively encodes group invariance, matching the performance of the grayscale oracle ($72.8\%$ vs. $73.1\%$).
Critically, because the invariance is defined by explicit pairs rather than dataset labels, the method is resilient to the inherent label noise (Appendix~\ref{app:ablation}).
This suggests that for simple spurious correlations, explicit invariance guidance allows the model to discard spurious information as effectively as preprocessing steps that remove it.

\paragraph{Extrapolation and Semantic Agnosticism.}
On Waterbirds-100, IPG achieves state-of-the-art results (86.3\% worst-group accuracy), outperforming baselines that rely on other guiding mechanisms using standard additive regularization. 
This confirms that sparse invariance pairs can successfully guide the model to extrapolate to unseen groups (e.g., waterbirds on land) even when the training set provides zero support.
Furthermore, the equally strong performance in the reversed setting demonstrates that IPG is agnostic to the semantic nature of the artifact.
Whether the spurious feature is a background texture or a foreground object, the method mitigates reliance on the shortcut once it is defined by the pairs.

\paragraph{Real-World Entanglement and Pair Quality.}
On CelebA, IPG acts as a \emph{conservative regularizer}: improves worst-group accuracy by 26.9 percentage points over ERM while maintaining superior mean test accuracy compared to implicit methods like JTT. However, the remaining gap to dense performance supervision highlights the challenge of \textit{semantic entanglement}.

Unlike Waterbirds, real-world attributes like gender and hair style are coupled; altering one often necessitates changing the other to maintain photorealism~\cite{pan_counterfactual_2018}. This violates the assumption of independent intervention, resulting in generated pairs with inevitable artifacts.

Furthermore, generating these counterfactuals synthetically introduces a distribution shift. 
Instances within a pair thus differ by both the spurious feature and this shift, corrupting the correction signal. 
Although IPG is sensitive to semantic mismatches, it can still substantially mitigate spurious correlations, a robustness demonstrated via our pair-building ablations (Appendix~\ref{app:pair-quality}). 
Ultimately, because semantic noise bottlenecks peak performance, applying IPG to highly entangled domains requires carefully curated pairs to minimize distribution shifts and entanglement-induced errors.

\paragraph{Theoretical Integrity.}
The efficacy of IPG is supported by our proof of monotonic descent toward the invariant region.
Specifically, the corrective gradient mathematically forces monotonic descent toward the invariant region, minimizing alignment error until the invariance condition is satisfied. 
However, because general neural networks violate our assumptions of global convexity and $L$-smoothness, the practical applicability of these convergence guarantees is generally limited to the vicinity of local minima. 
Future work will formally analyze the \textit{stability} of this invariant region, particularly the conditions under which the task gradient might displace parameters once the corrective force is reduced (Appendix~\ref{app:stability}).

\paragraph{Complexity, Efficiency, and Limitations.}
Although our experiments used a configuration ($B_I=B_D$) resulting in a $3\times$ overhead, Appendix~\ref{app:sensitivity} suggests a reduction to an overhead of $\approx 1.5\times$ ERM results in comparable accuracy.
Our data-efficiency results show that robust features can be recovered with 124 pairs.
However, certain limitations remain. 
First, performance is bounded by the quality of invariance pairs; poor generation can misguide the model, though improvements in generative AI will likely ease this bottleneck. 
Second, the current rationale formulation is specific to classification.

\section{Conclusion}
\label{sec:conclusion}

In this work, we presented \textit{Invariance Pair Guidance (IPG)}, a framework that decouples robust learning from the need for dense supervision or bias-conflicting data.
By leveraging a sparse set of invariance pairs, IPG formulates a corrective gradient that penalizes model rationales only when they diverge from the desired invariance.
Empirically, IPG demonstrates remarkable universality: it matches oracle performance on synthetic data (ColoredMNIST), achieves state-of-the-art extrapolation on data-scarce regimes (Waterbirds-100), and acts as a conservative regularizer on real-world data (CelebA).
Our theoretical analysis supports these findings, guaranteeing monotonic descent toward the invariant region under standard assumptions.
Crucially, IPG offers a data-efficient path to robustness that eliminates the 
need for dataset preprocessing or multi-stage training pipelines.
In future, we will focus on integrating IPG with latent-space generative models to synthesize high-fidelity, disentangled pairs for increasingly complex real-world attributes.
In addition, we plan to generalize the framework to multi-objective settings.

\bibliography{references}

\newpage
\begin{appendices}

\section{Detailed Sensitivity Analysis}
\label{app:sensitivity}

In this section, we provide the full empirical results regarding hyperparameter sensitivity and ablation studies, complementing the summary in Section~\ref{sec:effectiveness_sensitivity}.

\subsection{Hyperparameter Sensitivity}
We analyzed the impact of the corrective gradient step size $\stepSize$ and the disagreement threshold $\thresh$ across 10 independent trials.
Table~\ref{tab:app_sensitivity_cmnist} and Table~\ref{tab:app_sensitivity_real} visualize the trade-off between strict invariance enforcement and task learning.

\paragraph{Synthetic vs. Real-World Dynamics.}
On ColoredMNIST (Table~\ref{tab:app_sensitivity_cmnist}) and Waterbirds-100 (Table~\ref{tab:app_sensitivity_real}), the best performance is achieved with strict constraints ($\thresh=0.1, \stepSize=0.1$). 
Since the invariance pairs clearly define the spurious correlation, any deviation from the invariance condition indicates reliance on spurious correlation, requiring immediate correction.
Conversely, on CelebA (Table~\ref{tab:app_sensitivity_real}), the model benefits from a looser threshold ($\thresh \approx 0.5$).
The value allows the optimization to suppress minor inconsistencies in the rationales of the invariance pairs, preventing the corrective gradient from overriding useful semantic feature learning.

As a heuristic for new datasets, we propose default starting values of $\alpha=0.1$ and $\tau=0.0$ to reduce the need for iterative retraining. 
Based on empirical experience, $\alpha$ only requires tuning (increasing) when the primary learning signal is weak, while $\tau$ serves as a tolerance threshold that should only be increased if the pair formulation can contain misleading pairs.

\begin{table}[h]
    \centering
    \caption{Hyperparameter sensitivity of $\accWg$ on ColoredMNIST. Strict thresholds ($\thresh=0.1$) yield optimal results by enforcing total invariance to color.}
    \label{tab:app_sensitivity_cmnist}
    \begin{tabular}{
    c| 
    S[separate-uncertainty=true, table-format=2.1, table-figures-uncertainty=3] 
    S[separate-uncertainty=true, table-format=2.1, table-figures-uncertainty=3] 
    S[separate-uncertainty=true, table-format=2.1, table-figures-uncertainty=3] 
}
    \toprule
    & \multicolumn{3}{c}{ColoredMNIST ($\thresh$)} \\
    $\stepSize$ & {0.1} & {0.5} & {0.9} \\
    \midrule
    0.1 & 68.9 +- 9.0 & 33.2 +- 6.6 & 15.3 +- 2.0\\
    0.5 & 59.9 +- 14.5 & 32.5 +- 4.2 & 14.8 +- 2.4\\
    0.9 & 48.7 +- 15.5 & 32.3 +- 3.6 & 14.5 +- 1.3\\
    \bottomrule
\end{tabular}
\end{table}

\begin{table}[h]
    \centering
    \caption{Hyperparameter sensitivity of $\accWg$ on Waterbirds-100 and CelebA.}
    \label{tab:app_sensitivity_real}
\begin{tabular}{
    c|
    S[separate-uncertainty=true, table-format=2.1, table-figures-uncertainty=1]
    S[separate-uncertainty=true, table-format=2.1, table-figures-uncertainty=1]
    S[separate-uncertainty=true, table-format=2.1, table-figures-uncertainty=1]|
    S[separate-uncertainty=true, table-format=2.1, table-figures-uncertainty=1]
    S[separate-uncertainty=true, table-format=2.1, table-figures-uncertainty=1]
    S[separate-uncertainty=true, table-format=2.1, table-figures-uncertainty=1]
}
    \toprule
    & \multicolumn{3}{c|}{Waterbirds-100 ($\thresh$)} & \multicolumn{3}{c}{CelebA ($\thresh$)} \\
    $\stepSize$ & {0.1} & {0.5} & {0.9} & {0.1} & {0.5} & {0.9} \\
    \midrule
    0.1 & 81.0 +- 1.9 & 78.2 +- 0.8 & 78.4 +- 1.1 & 68.9 +- 5.2 & 68.4 +- 4.0 & 71.7 +- 4.3 \\
    0.5 & 80.9 +- 1.9 & 78.2 +- 0.9 & 78.0 +- 1.0 & 69.8 +- 5.2 & 71.1 +- 8.4 & 71.3 +- 6.4 \\
    0.9 & 79.7 +- 1.3 & 78.3 +- 0.8 & 78.3 +- 0.8 & 70.9 +- 7.9 & 74.5 +- 5.7 & 70.9 +- 3.7 \\
    \bottomrule
\end{tabular}
\end{table}

\paragraph{Conservative upper bound $B_I=B_D$.}
Adjusting the ratio between the invariance batch size ($B_I$) and the dataset batch size ($B_D$) offers a practical approach to improving computational runtime. 
Although we set $B_I=B_D=128$ in our primary experiments, here we evaluate the sensitivity of the model's accuracy to smaller invariance batch sizes, specifically $B_I \in \{32, 64, 128\}$. 
As detailed in Table~\ref{tab:control-batch-size-sensitivity}, the impact on performance is marginal: we observe a maximal change of $-0.3$ percentage points for ColoredMNIST, $+0.3$ percentage points for Waterbirds-100, and $-0.1$ percentage points for CelebA. 
These results demonstrate that the accuracy exhibits limited sensitivity to $B_I$, even though decreasing $B_I$ accelerates the correction step by a factor proportional to $B_I/B_D$. 
Therefore, setting $B_I=B_D$ serves as a conservative upper bound.

\begin{table}[htb]
    \centering
    \caption{
    Sensitivity of $\accWg$ to the value of $B_I$.
    }
    \label{tab:control-batch-size-sensitivity}

    \begin{tabular}{
        c S[table-format=1.2] |
        S[table-format=1.3(3)] 
        S[table-format=1.3(3)] 
        S[table-format=1.3(3)] 
    }
        \toprule
        {$B_I$} & {$B_I/B_D$} & {ColoredMNIST} & {Waterbirds-100} & {CelebA} \\
        \midrule
        32  & 0.25 & \num{72.0 \pm 1.8} & {\bfseries\num{86.6 \pm 1.0}} & \num{74.6 \pm 5.4} \\
        64  & 0.5 & \num{72.8 \pm 0.5} & \num{86.3 \pm 1.2} & \num{74.6 \pm 6.0} \\
        128 & 1.0 & {\bfseries\num{72.8 \pm 0.5}} & \num{86.3 \pm 1.9} & {\bfseries\num{74.7 \pm 8.1}} \\

        \bottomrule
    \end{tabular}
\end{table}

\subsection{Data Efficiency Visualization}
\begin{figure}[h]
    \centering
    \includegraphics[width=.8\textwidth]{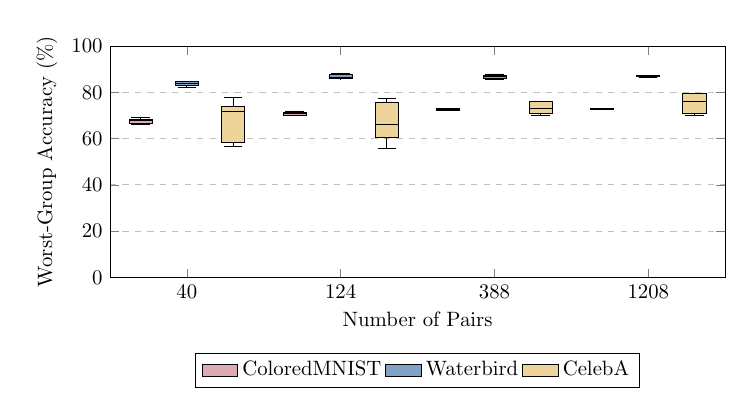}
    \caption{Distribution of worst-group accuracy over 10 trials with varying number of invariance pairs.}
    \label{fig:app_nr_pairs}
\end{figure}

In the main text, we reported the mean performance across varying numbers of pairs. 
Figure~\ref{fig:app_nr_pairs} provides the corresponding boxplot distribution, highlighting the variance reduction as the number of pairs increases.

\subsection{Ablation Studies}
\label{app:ablation}

To validate the specific design choices of IPG, we compare our method against two simplifications of the optimization mechanism (Table~\ref{tab:app_ablation}):

\begin{enumerate}
    \item \textbf{Joint Optimization (Sum of Losses):} 
    A standard multi-task formulation where the task loss $\lossTask$ and the alignment loss $\lossAlign$ are applied consecutively. 
    This removes the adaptive gradient scaling.
    \item \textbf{Continuous Correction (Full Control):} 
    A variant of IPG where the invariance threshold is set to an unreachable strict limit ($\thresh = 0$). 
    In this scenario, the invariance condition is effectively never satisfied, forcing the gradient correction to be applied at every single step. 
    This tests whether the correction relaxation of our correction for the invariant region is beneficial.
\end{enumerate}

As shown in Table~\ref{tab:app_ablation}, Joint Optimization fails to recover the Oracle performance on ColoredMNIST ($10.2\%$ vs. $72.8\%$) or the performance on Waterbirds-100 ($33.1\%$ vs. $86.3\%$).
The difference indicates that simply adding up the losses is not sufficient to control for the spurious correlation.
The model fails to prioritize the invariance over the strong spurious signal.

Continuous Correction achieves strong results on ColoredMNIST ($72.8\%$) and Waterbirds-100 ($86.3\%$), which is expected since the spurious correlation is perfect and requires constant suppression. 
However, on real-world datasets like CelebA, continuous correction degrades performance slightly compared to IPG ($70.9\%$ vs. $74.7\%$). 
This indicates that the model benefits from an relaxed invariance condition allowing a certain rate of mismatches. 

\begin{table}[h]
    \centering
    \caption{
    Ablation study comparing IPG with baselines on $\accWg$. 
    }
    \label{tab:app_ablation}
    \begin{tabular}{l c c c}
        \toprule
        Method & ColoredMNIST & Waterbirds-100 & CelebA \\
        \midrule
        Joint Optimization & \num{10.2 \pm 0.4} & \num{33.1 \pm 4.0} & \num{72.2 \pm 5.2} \\
        Continuous Correction ($\thresh = 0$) & \num{72.8 \pm 0.5} & \num{86.1 \pm 2.0} & \num{70.9 \pm 9.0} \\
        \textbf{IPG} & \textbf{\num{72.8 \pm 0.5}} & \textbf{\num{86.3 \pm 1.9}} & \textbf{\num{74.7 \pm 8.8}} \\
        \bottomrule
    \end{tabular}
\end{table}

In addition, we investigate the effect of label noise on IPG.
We compare the performance of IPG on two variants of the ColoredMNIST dataset \cite{arjovsky_invariant_2019}.
First, we consider a variant without any label noise. 
Second, we consider a variant which applies label noise with probability $p_\text{noise,train} = 0.25$ to samples in the train and validation splits after the color determination and dataset splitting.
In both cases, the test set is free of label noise.
Therefore, both variants have the same theoretical optimum of 100\%, which allows us to compare the results.
Table~\ref{tab:label_noise_ablation} shows the results. 
The results of the training without noisy labels are slightly (+2.9 percentage points) better and could be explained by the noise-free training signal. 
In both cases, the spurious correlation was effectively mitigated. 

\begin{table}[h]

    \centering
    \caption{
    Ablation study comparing IPG on ColoredMNIST variants. 
    }
    \label{tab:label_noise_ablation}
    \begin{tabular}{l c }
        \toprule
        Dataset & Mean Acc. \\
        \midrule
        ColoredMNIST ($p_\text{noise} = 0$) & \textbf{\num{97.5 \pm 0.6}} \\
        ColoredMNIST ($p_\text{noise,train} = 0.25$) & \num{94.6 \pm 4.1} \\
        \bottomrule
    \end{tabular}
\end{table}

\newpage
\section{Implementation Details}
\label{app:implementation}

\subsection{Training and Model Selection}
We train the models with the following specifications.
For ColoredMNIST, we employ a standard Convolutional Neural Network (CNN) following the architecture defined in the DomainBed framework~\cite{gulrajani_search_2021}.
For Waterbirds-100 and CelebA, we use a ResNet-50 backbone~\cite{he_deep_2016} pretrained on ImageNet.
All models are optimized using Adam with a class-weighted cross-entropy loss. 
All results are evaluated over $10$ independent trials with different random seeds, including the mean and standard deviation of the accuracy values. 
For ColoredMNIST, model selection is based on the highest mean validation accuracy within the training environments ($p_\text{tr} \in \{0.1, 0.2\}$). 
To prevent data leakage during hyperparameter optimization, we partitioned a dedicated hold-out validation set from the unseen test environment ($p_\text{te} = 0.9$). %
For Waterbirds-100 and CelebA, model selection is evaluated based on the maximum worst-group validation accuracy using their standard official splits.

\subsection{Hyperparameters}
The hyperparameters for our approach are listed in Table~\ref{tab:hyper-params} and are selected using the Tree-structured Parzen Estimator with $100$ trials per dataset \cite{bergstra_algorithms_2011}. 
We apply the same batch size for data and invariance pair batches.
Hyperparameter optimization is performed strictly using the official validation splits for Waterbirds-100 and CelebA. 
For ColoredMNIST, optimization is conducted exclusively on the newly introduced hold-out validation split to ensure strict separation from the final test evaluation.

\begin{table}[ht]
\centering
\caption{Hyperparameter configuration of IPG in experiments.}
\label{tab:hyper-params}
\begin{tabular}{l l
    S[table-format=1.2]
    S[table-format=1.2]
    S[table-format=1.1]
    S[table-format=3.0]
    S[table-format=1.1e-1]
    S[table-format=3.0]
    S[table-format=2.0]
}

    \toprule
    Dataset & Model & {$\stepSize$} & {$\thresh$} & {$\margin$} & {$|\invpairs|$} & {$\lr$} & {$B_D$} & {Nr. Epochs} \\
    \midrule
    ColoredMNIST & IPG & \num{0.5} & \num{0.0} & \num{0.08} & \num{1208} & \num{1e-3} & \num{128} & \num{18} \\
    Waterbirds-100 & IPG & \num{0.1} & \num{0.0} & \num{0.1} & \num{124} & \num{1e-4} & \num{128} & \num{10} \\
    Waterbirds-100-reverse & IPG & \num{0.5} & \num{0.5} & \num{0.1} & \num{124} & \num{1e-4} & \num{128} & \num{10} \\
    CelebA & IPG & \num{0.5} & \num{1e-3} & \num{1.5} & \num{1208} & \num{8e-3} & \num{128} & \num{10} \\
    \bottomrule
\end{tabular}
\end{table}

\newpage
\section{Theoretical Guarantees and Training Dynamics}
\subsection{Proof of Monotonic Correction and Convergence Properties}
\label{app:proof}

We provide the full proof of the Theorem~\ref{thm:monotonic_desc} and Corollary~\ref{cor:convergence_properties}.

\monotonicDescent*
\begin{proof}
Since $\disagree \ge \thresh$, the update rule (Alg. \ref{alg:approach}) applies a correction. 
The task gradient is scaled such that:
\begin{equation}\|\gradTask\| = \stepSize \|\gradCorr\|.\label{eq:convergence_grad_correction}\end{equation}
The parameter update is given by:
\begin{equation*}\label{eq:update}
\params_{k+1} = \params_k -\lr (\gradCorr+\gradTask).
\end{equation*}
Using the $L$-smoothness upper bound (Assumption \ref{ass:smoothness}):\begin{equation}\label{eq:conv_smooth_inequality}
\lossCorrParams{k+1} \leq \lossCorrParams{k} \underbrace{-\lr \nabla \lossCorrParams{k}^T(\gradCorr+\gradTask)}_{\text{Linear Term}} + \underbrace{\frac{L}{2}\lr^2\|\gradCorr+\gradTask\|^2}_{\text{Quadratic Term}}.\end{equation}
We analyze the linear term. 
In the worst-case scenario, the task gradient $\gradTask$ opposes the corrective gradient. 
Using the Cauchy-Schwarz inequality and Eq. \ref{eq:convergence_grad_correction}:\begin{equation}\label{eq:linear_term}
\begin{split}
-\lr \gradCorr^T (\gradCorr + \gradTask) 
&= -\lr \|\gradCorr\|^2 - \lr \langle \gradCorr, \gradTask \rangle \\
&\leq -\lr \|\gradCorr\|^2 + \lr \|\gradCorr\| \|\gradTask\| \\
&= -\lr (1 - \stepSize) \|\gradCorr\|^2.
\end{split}
\end{equation}
This term is negative (providing descent) only if $\stepSize < 1$.
Next, we bound the quadratic penalty term using the triangle inequality:\begin{equation}\label{eq:quad_term}
\begin{split}
\|\gradCorr + \gradTask\|^2 
&\leq (\|\gradCorr\| + \|\gradTask\|)^2 \\
&= (1 + \stepSize)^2 \|\gradCorr\|^2.
\end{split}
\end{equation}
Substituting Eqs. \ref{eq:linear_term} and \ref{eq:quad_term} back into Eq. \ref{eq:conv_smooth_inequality}:\begin{equation*}\label{eq:combined_inequality}
\lossCorrParams{k+1} \leq \lossCorrParams{k} - \lr \left[ (1 - \stepSize) - \frac{L\lr}{2}(1 + \stepSize)^2 \right] \|\gradCorr\|^2.
\end{equation*}
For monotonic descent ($\lossCorrParams{k+1} < \lossCorrParams{k}$), the term in the brackets must be positive.
Since $\stepSize < 1$, the term $(1-\stepSize)$ is strictly positive. 
Therefore, there exists a sufficiently small learning rate $\lr$ such that:
\begin{equation*}
\frac{L\lr}{2}(1 + \stepSize)^2 < 1 - \stepSize.
\end{equation*}
Under this condition, the loss strictly decreases.
\end{proof}

\convergenceProperties*
\begin{proof}
Let $\gamma$ denote the effective stopping threshold. For the relaxed case, $\gamma = \tau$. 
For the strict case ($\tau=0$), $\gamma = \epsilon$, where $\epsilon > 0$ is the stability margin of the discrete classifier. 
Since predictions are equal to aggregating rationales (Eq.~\ref{eq:rationale_prediction}), the local Lipschitz continuity ensures that small deviations in $\mathbf{R}$ do not alter class labels provided the error is within this margin.
Thus, in both cases, the target condition is satisfied if the continuous alignment error falls below $\gamma$.
We prove that the algorithm satisfies this condition in finite time by contradiction. 
Assume the optimization never satisfies the condition, meaning the alignment error remains above $\gamma$ for all $k$. 
Under this assumption, the corrective update is applied indefinitely.
By Theorem~\ref{thm:monotonic_desc} (Monotonic Descent), the loss sequence $\{\lossCorrParams{k}\}$ is strictly decreasing and bounded below by 0. 
By the Monotone Convergence Theorem and Assumption~\ref{ass:joint_feasibility} (Feasibility), the sequence converges to the global minimum where the alignment error is zero:
\begin{equation*}
\lim_{k \to \infty} \lossAlignSub{, \params_k} = 0.
\end{equation*}
By the definition of the limit, for any $\gamma > 0$, there exists a finite iteration $K$ such that for all $k \geq K$, $\lossAlignSub{, \params_k} < \gamma$.
This contradicts the assumption that the error remains above $\gamma$ indefinitely.
Therefore, the process must reach the target state (entering $\invRegion$ for $\tau > 0$, or achieving exact $\disagree = 0$ for $\tau = 0$) in a finite number of steps.
\end{proof}

\subsection{Training Dynamics on ColoredMNIST}
\label{app:full-batch}
\begin{figure}[htb]
    \centering
    \includegraphics[width=0.8\textwidth]{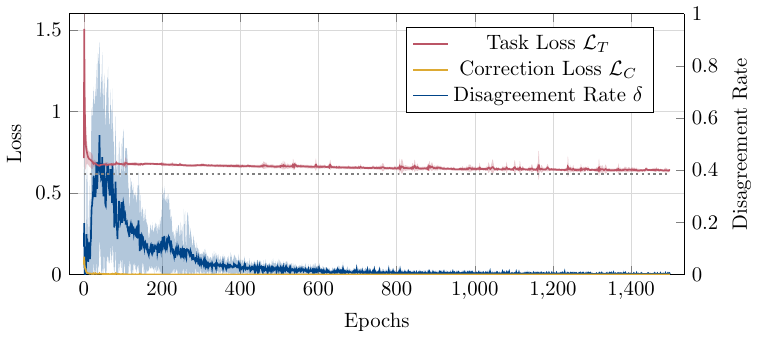}
    \caption{
    Loss evolution during full-batch training on ColoredMNIST.
    Solid lines and shaded areas represent the mean and standard deviation across 10 independent trials, respectively.
    The dotted line highlights the minimum observed task loss ($\approx 0.6$). 
    }
    \label{fig:full-batch}
\end{figure}

To complement the theoretical insights, we visualize the dynamics during training. 
We conduct 10 trials of full-batch training on the ColoredMNIST dataset with specific hyperparameters ($\eta = $ \num{5e-3}, \num{1500} epochs, $\alpha = 0.1$). 
Figure~\ref{fig:full-batch} illustrates the evolution of $\lossTask$, $\lossCorr$, and $\disagree$. 
Remarkably, $\lossTask$ exhibits a strict lower bound of $\approx 0.6$ after a sharp spike in the first epoch.
Furthermore, $\disagree$ and $\lossCorr$ peak during the first 100 epochs and subsequently decay to 0.
These results indicate that after a correction phase of 400 epochs, spurious correlation is effectively mitigated. 
In addition, the task loss is strictly constrained by the correction mechanism throughout the training.

\subsection{Remarks on the Stability of the Method}
\label{app:stability}

Once the parameters enter a region $\invRegion$, where the invariance condition is satisfied (i.e., $\disagree < \thresh$), the standard task gradient may exceed the corrective gradient in magnitude (e.g., $\|\nabla_{\params} \lossTask\|_2 \gg \|\gradCorr\|_2$). 
While minor displacements from this invariant state are expected under such conditions, the task gradient must be prevented from causing severe divergence that would require a prohibitive number of corrective steps to overcome. 
To bound the extent of these displacements, the two gradients must be dynamically coupled, a process we refer to as gradient harmonization. 

As defined in the ``else'' clause of Eq.~\ref{eq:loss_grad}, we achieve this by bounding the magnitude of $\gradTask$ to $2\cdot\|\gradCorr\|_2$. 
While the corrective gradient generally maintains a non-zero magnitude in this region for datasets with strong spurious correlations, in the event that $\|\gradCorr\|_2$ approaches zero, strictly bounding the task gradient by this value would cause gradient starvation. 
To prevent this, we enforce a strict lower bound $\varepsilon$, effectively clipping the task gradient to a maximum length of $2\cdot \max(\|\gradCorr\|_2, \varepsilon)$. 
Because this dynamic bounding strictly limits the maximum possible divergence per step, the corrective gradient reliably regains dominance whenever the invariance condition is violated. 
This ensures that the system is continually driven back toward the invariant state (with the step magnitude controlled by $\alpha$), consistent with the monotonic convergence properties established in Theorem~\ref{thm:monotonic_desc}.
While a formal stability analysis of this invariant region is left for future work, this continuous cycle of monotonic convergence and dynamic dampening of divergent forces is a hallmark of stable dynamical systems. 

\section{Pair Generation}
\label{app:pair-generation}
This section details the generation process for the invariance pairs, providing visual examples, the specific text prompts employed for the CelebA dataset, and an analysis of common failure modes.

\subsection{Pair Illustrations}
\begin{figure}[htb]
    \centering
    \includegraphics[width=0.45\linewidth]{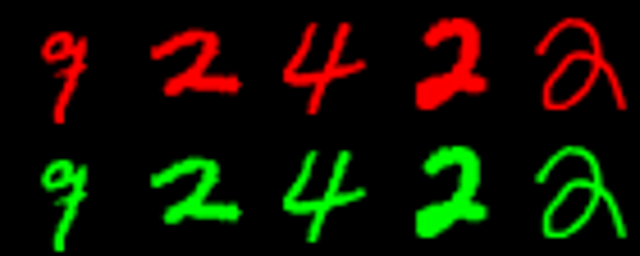}
    \caption{Representative invariance pairs for the ColoredMNIST dataset.}
    \label{fig:pairs-coloredmnist}
\end{figure}

\begin{figure}[htb]
    \centering
    \includegraphics[width=0.6\linewidth]{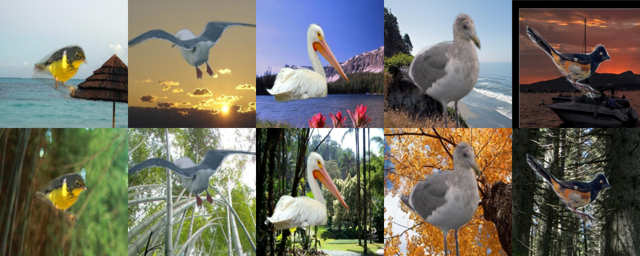}
    \caption{Representative invariance pairs for Waterbirds-100.}
    \label{fig:pairs-waterbird}
\end{figure}

\begin{figure}[htb]
    \centering
    \includegraphics[width=0.6\linewidth]{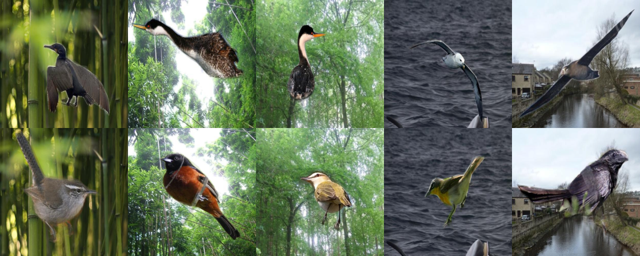}
    \caption{Representative invariance pairs for Waterbirds-100-reverse.}
    \label{fig:pairs-waterbird-reverse}
\end{figure}

\begin{figure}[htb]
    \centering
    \includegraphics[width=0.6\linewidth]{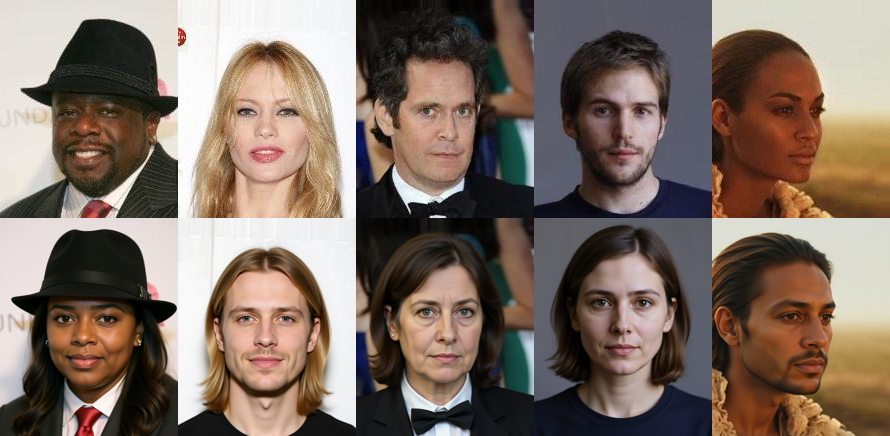}
    \caption{Representative invariance pairs for CelebA. Training images in the first row, generated images in the second row.}
    \label{fig:pairs-celeba}
\end{figure}

We provide visual representations of the invariance pairs generated for the three benchmark datasets: ColoredMNIST (Fig.~\ref{fig:pairs-coloredmnist}), Waterbirds-100 (Figs.~\ref{fig:pairs-waterbird} and \ref{fig:pairs-waterbird-reverse}), and CelebA (Fig.~\ref{fig:pairs-celeba}). 
These examples demonstrate the isolation of spurious attributes while retaining semantic content.

\subsection{Prompts for CelebA Pair Generation}
\label{app:prompts}
To generate counterfactual pairs (Fig.~\ref{fig:pairs-celeba})---specifically, gender-transformed images---we utilized Flux.1 [Context] dev \cite{labs_flux_2025}, a rectified flow transformer capable of editing images based on text instructions. 

For this process, we randomly sampled $1,300$ instances from each group in the training set. 
For each source image, three candidate counterfactuals were generated using specific textual prompts. Below, we provide examples of the prompts used.

For instance, when transforming a source image of a dark-haired male, the following prompt was used:

\lstset{
  basicstyle=\ttfamily\small,
  showstringspaces=false,
  keywords={female, male},
  keywordstyle=\color[HTML]{004488}
}
\begin{lstlisting}
Transform this male person into a female version, retaining the 
unique characteristics, especially hair color, age, and position.
The result is photorealistic, with natural lighting.
\end{lstlisting}

Conversely, for samples involving blond hair (e.g., transforming a blond female), the target hair color was explicitly specified to prevent attribute leakage:

\lstset{
  basicstyle=\ttfamily\small,
  showstringspaces=false,
  keywords={female, male},
  keywordstyle=\color[HTML]{004488}
}
\begin{lstlisting}
Transform this female person into a male version with clearly 
blond hair, retaining the unique characteristics, especially age
and position. The result is photorealistic, with natural 
lighting.
\end{lstlisting}

Prompts for other subgroups follow a similar template, with the gender-specific keywords (highlighted in blue) substituted accordingly.

\subsection{Evaluating the Impact of Pair Quality}
\label{app:pair-quality}

\begin{figure}[htb]
    \centering
    \begin{subfigure}[b]{0.45\textwidth}
        \includegraphics[width=\textwidth]{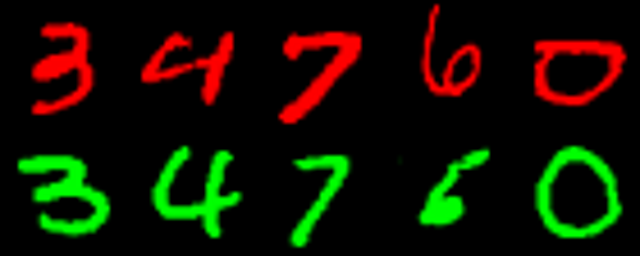}
        \caption{}
    \end{subfigure}
    \begin{subfigure}[b]{0.45\textwidth}
        \includegraphics[width=\textwidth]{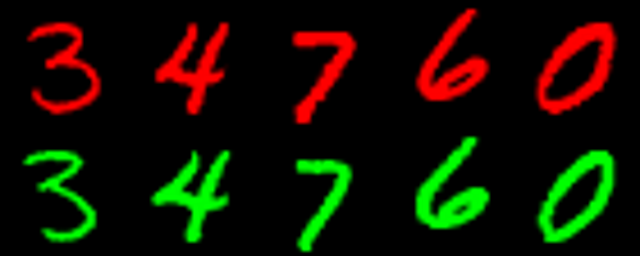}
        \caption{}
    \end{subfigure}
    \caption{Representative pairs of the (a) random and (b) closest pair selection strategy.}
    \label{fig:pair-strategy}
\end{figure}

Throughout the paper, we apply a \emph{perfect} pair-building strategy for ColoredMNIST by switching color-channels of the same digit.
We consider alternative approaches that lead to diminished pair quality, allowing us to investigate the sensitivity of IPG to pair quality.
To select the pairs for the experiment, we consider two additional strategies. 
The \emph{closest} strategy relies on the latent representation of a trained oracle model, which serves as an expert proxy.
Specifically, we train a model on the grayscale images via ERM.
For each digit, we select pairs of training instances that exhibit the shortest Euclidean distance withing the oracle's latent space.
Furthermore, the \emph{random} strategy randomly pairs instances of the same digit from the training split. 
Examples pairs generated by both strategies are illustrated in Figure~\ref{fig:pair-strategy}.

Because the \emph{closest} and \emph{random} strategies introduce a noisy signal due to the semantic mismatch of imperfect pairs, we decrease the step size $\alpha$ to 0.1 to strengthen the corrective mechanism.
All other hyperparameters remain as defined in Appendix~\ref{app:implementation}.
Table~\ref{tab:pair-strategy} provides an overview of the results.
We observe that for the embedding-based (\emph{closest}) selection strategy, the mean test accuracy drops slightly to 67.2\%.
By comparison, the mean accuracy for the \emph{random} selection strategy is 62.6\%.
While this indicates that the model achieves partial invariance, its accuracy is not comparable to that of the other strategies.
These results demonstrate a clear trade-off: while perfect pair construction yields near-optimal performance, our embedding-based strategy proves that robust invariances can still be learned without perfectly matched samples.

\begin{table}[h]

    \centering
    \caption{
    Accuracy comparison on ColoredMNIST with different pair-building strategies.
    }
    \label{tab:pair-strategy}
    \begin{tabularx}{0.65\textwidth}{
    >{\raggedright\arraybackslash}X
    >{\centering\arraybackslash}X
    >{\centering\arraybackslash}X
    }
        \toprule
        Strategy & $\alpha$ & Mean Acc. \\
        \midrule
        random & 0.1 & \num{62.6 \pm 2.3} \\
        closest & 0.1 & \num{67.2 \pm 1.0} \\
        perfect & 0.5 & \textbf{\num{72.8 \pm 0.5}} \\
        \bottomrule
    \end{tabularx}
\end{table}

\subsection{CelebA Failure Cases}
\label{app:celeba-failure-cases}
\begin{figure}
    \centering
    \includegraphics[width=0.6\linewidth]{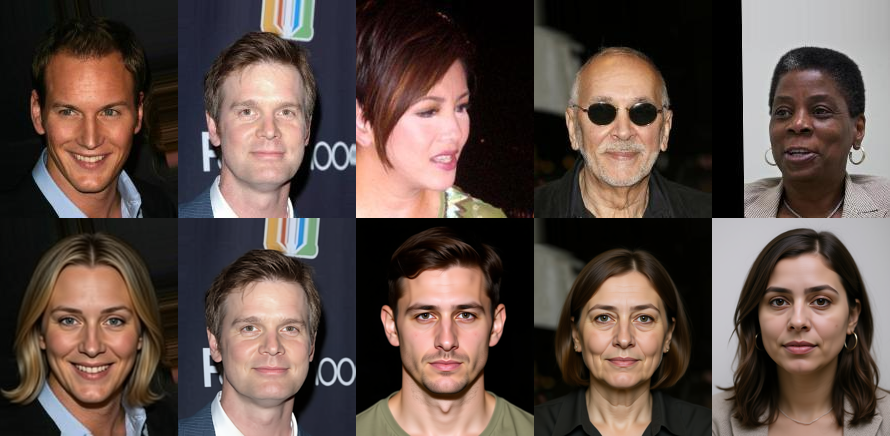}
    \caption{Examples of discarded pairs exhibiting generation artifacts or attribute inconsistencies. Training images in the first row, generated images in the second row.}
    \label{fig:celeba-failure-cases}
\end{figure}

Figure~\ref{fig:celeba-failure-cases} illustrates instances where the generation process failed to meet the invariance criteria. Common failure modes included: (i) unintentional alteration of the hair color, (ii) unsuccessful gender transformation, or (iii) significant modification of extraneous attributes such as accessories, skin tone, or head pose.

\end{appendices}
\end{document}

%% file: symbols.tex
\usepackage{bm}
\usepackage{bbm} %

\newcommand{\params}{\bm{\theta}}
\newcommand{\grad}{\mathbf{g}}
\newcommand{\R}{\mathbf{R}} %
\newcommand{\z}{\mathbf{z}}
\newcommand{\x}{\mathbf{x}}
\newcommand{\y}{\mathbf{y}}
\newcommand{\update}{\text{OptimizerStep}}
\newcommand{\domain}{\mathcal{D}} %
\newcommand{\loss}{\mathcal{L}} %
\newcommand{\inputspace}{\mathcal{X}} %
\newcommand{\invpairs}{\mathcal{I}} %
\newcommand{\invpairsbatch}{\mathbf{I}} %
\newcommand{\invpairpred}[1][]{\hat{\textbf{y}}_{\text{inv}#1}}

\newcommand{\gradCorr}{\grad_{C}}
\newcommand{\gradTask}{\grad_{T}}

\newcommand{\stepSize}{\alpha}

\newcommand{\thresh}{\tau}     %
\newcommand{\batchX}{\x}
\newcommand{\batchY}{\y}
\newcommand{\invX}{\x_{\text{inv}}}

\newcommand{\lr}{\eta}

\newcommand{\lossTask}{\mathcal{L}_T}
\newcommand{\lossCorr}{\mathcal{L}_C}
\newcommand{\lossCorrSub}[1]{\mathcal{L}_{C{#1}}}
\newcommand{\lossCorrParams}[1]{\lossCorrSub{,\params_{#1}}}
\newcommand{\lossAlign}{\mathcal{L}_A}
\newcommand{\lossUniform}{\mathcal{L}_U}
\newcommand{\lossAlignSub}[1]{\mathcal{L}_{A{#1}}}
\newcommand{\lossUniformSub}[1]{\mathcal{L}_{U{#1}}}
\newcommand{\margin}{m}
\newcommand{\ind}{\mathbbm{1}}
\newcommand{\softmax}{\text{softmax}}
\newcommand{\disagree}{\delta} %
\newcommand{\diag}{\text{diag}}
\newcommand{\extractor}{\phi}
\newcommand{\invRegion}{\mathcal{R}_\thresh}
\newcommand{\paramsDomain}{\bm{\Theta}}
\newcommand{\accWg}{\text{Acc}_\text{wg}}
\newcommand{\trainDomain}{\domain_\text{tr}}
\newcommand{\testDomain}{\domain_\text{te}}